%% file: main.tex
\documentclass{article}

\usepackage{microtype}
\usepackage{graphicx}
\usepackage{booktabs} 
\usepackage{algorithm}
\usepackage{algpseudocode}
\usepackage{subcaption}
\usepackage{hyperref}
\usepackage[dvipsnames]{xcolor}
\usepackage{amsfonts,amsmath,amssymb}

\usepackage[accepted]{mlsys2026}

\input{math_commands.tex}

\newcommand{\methodname}{CAGE}

\renewcommand{\paragraph}[1]{\noindent\textbf{#1}}

\mlsystitlerunning{\methodname: \textbf{C}urvature-\textbf{A}ware \textbf{G}radient \textbf{E}stimation For Accurate Quantization-Aware Training}

\newtheorem{lemma}[theorem]{Lemma}

\begin{document}

\twocolumn[
\mlsystitle{\methodname: \textbf{C}urvature-\textbf{A}ware \textbf{G}radient \textbf{E}stimation For Accurate Quantization-Aware Training}



\mlsyssetsymbol{equal}{*}

\begin{mlsysauthorlist}
\mlsysauthor{Soroush Tabesh}{equal,ista}
\mlsysauthor{Mher Safaryan}{equal,ista}
\mlsysauthor{Andrei Panferov}{ista}
\mlsysauthor{Alexandra Volkova}{ista}
\mlsysauthor{Dan Alistarh}{ista,redhat}
\end{mlsysauthorlist}

\mlsysaffiliation{ista}{ISTA}
\mlsysaffiliation{redhat}{Red Hat AI}

\mlsyscorrespondingauthor{Dan Alistarh}{dan.alistarh@ista.ac.at}
\mlsyscorrespondingauthor{Soroush Tabesh}{soroush.tabesh@ista.ac.at}

\mlsyskeywords{Machine Learning, Quantization, QAT, Quantization-Aware Training, MLSys}

\vskip 0.3in

\begin{abstract}
Despite significant work on low-bit quantization-aware training (QAT), there is still an accuracy gap between such techniques and native training. To address this, we introduce CAGE (Curvature-Aware Gradient Estimation), a new QAT method that augments the straight-through estimator (STE) gradient with a curvature-aware correction designed to counteract the loss increase induced by quantization. CAGE is derived from a multi-objective view of QAT that balances loss minimization with the quantization constraints, yielding a principled correction term that depends on local curvature information.  On the theoretical side, we introduce the notion of Pareto-optimal solutions for quantized optimization, and establish that CAGE yields strong convergence guarantees in the smooth non-convex setting.
In terms of implementation, our approach is optimizer-agnostic, but we provide a highly-efficient implementation that leverages Adam statistics. 
CAGE significantly improves upon the prior state-of-the-art methods in terms of accuracy, for similar computational cost: for QAT fine-tuning, it halves the compression accuracy loss relative to the prior best method, while for QAT pre-training of Llama models, its accuracy for 3-bit weights-and-activations (W3A3) matches the accuracy achieved at \emph{4-bits} (W4A4) with the prior best method. The official implementation can be found over \href{https://github.com/IST-DASLab/CAGE}{\texttt{https://github.com/IST-DASLab/CAGE}}.
\end{abstract}
]



\printAffiliationsAndNotice{\mlsysEqualContribution} 

\section{Introduction}
\label{sec:intro}
\input{introduction}

\section{Prior Work}
\label{sec:related}
\input{related}


\section{\methodname{}: Curvature-Aware Gradient Estimation}
\label{sec:method}
\input{method}

\section{Experiments}
\label{sec:experiments}

\input{experiments}

\section{Summary and Discussion}
\label{sec:discussion}
\input{discussion}

\bibliography{references}
\bibliographystyle{mlsys2025}

\clearpage

\onecolumn
\appendix
\newpage 
\include{appendix}

\end{document}

%% file: math_commands.tex
\usepackage{amsmath,amsfonts,mathtools,bm}
\usepackage{nicefrac}

\newtheorem{theorem}{Theorem}

\newtheorem{assumption}{Assumption}

\newcommand{\eqdef}{\coloneqq} 

\def\eqref#1{(\ref{#1})}

\def\floor#1{\lfloor #1 \rfloor}
\def\1{\bm{1}}

\DeclareMathAlphabet{\mathsfit}{\encodingdefault}{\sfdefault}{m}{sl}
\SetMathAlphabet{\mathsfit}{bold}{\encodingdefault}{\sfdefault}{bx}{n}

\newcommand{\E}{\mathbb{E}}

\newcommand{\R}{\mathbb{R}}



%% file: introduction.tex
Quantization has emerged as a standard technique for improving the computational efficiency of large language model (LLM) deployments, as prominent open-source models such as Llama, Gemma, and GPT-OSS are released in compressed formats~\cite{touvron2023llama2openfoundation, team2024gemma, openai:gptoss}. 
Yet, the vast majority of open research in this area has concentrated on \emph{post-training quantization (PTQ)}, where a fully trained model's weights are quantized by applying numerical algorithms over a small calibration dataset~\cite{frantar2022gptq}. 
 
Relatively less is known about quantization-aware training (QAT)~\cite{bengio2013estimating, krishnamoorthi2018quantizing, esser2020lsq}, where the model learns to adapt to the constraints of quantization during the optimization process itself. QAT is more computationally-expensive, but can yield better accuracy. 
The state of the art for QAT methods has long been dominated by the straight-through estimator (STE), a method first suggested by~\citet{Hinton2012} and formalized by~\citet{bengio2013ste}, which is integrated by default in deep learning frameworks like PyTorch~\cite{pytorch}. This approach bypasses the non-differentiable quantization function in the backward pass, by approximating its gradient with the identity. While versatile, STE is known to suffer from slow convergence and instability. A body of work, such as LSQ~\citep{esser2020lsq}, LSQ+~\citep{bhalgat2020lsqplus}, and more recent methods~\cite{esser2019learned, bhalgat2020lsqplus, nagel2022overcoming, lee2021network}, has introduced improved approaches to stabilize and accelerate this process learnable quantization parameters or more sophisticated gradient scaling. Yet, these methods are heuristic, and do not offer formal convergence guarantees. 
 
\paragraph{Contributions.} 
In this paper, we address this gap by examining QAT from both a theoretical and a practical standpoint. We provide a formal framework for QAT as a multi-objective optimization problem, where the goal is to simultaneously minimize the task loss and the quantization error. From this perspective, we identify a condition for Pareto-optimality, where any improvement in one component of the objective necessarily leads to a degradation in the other. This formulation inspires a family of algorithms, called \methodname{}, for \textbf{C}urvature-\textbf{A}ware \textbf{G}radient \textbf{E}stimation, which provides both theoretical guarantees, and high accuracy and efficiency in practice. More precisely, our contributions are as follows: 

\begin{itemize}
\setlength\itemsep{0.5em}
\vspace{-0.3em}
\item  The \methodname{} method we introduce provides a principled method for augmenting the standard STE gradient with a curvature-aware correction term, derived directly from our Pareto-optimality condition. This term explicitly counteracts the increase in loss induced by the quantization step, by leveraging local second-order information about the loss landscape (i.e., the Hessian). 

\item 
From the theoretical perspective, we prove, under assumptions, that \methodname{} possesses strong ergodic convergence guarantees to a Pareto-optimal point in the smooth non-convex setting, directly addressing the lack of guarantees that is a primary limitation of prior work in this area. 
Our approach is optimizer-agnostic; we provide a highly-efficient implementation of this framework that utilizes statistics readily available in adaptive optimizers such as Adam. 

\item 
We validate the effectiveness of \methodname{} through extensive experiments, including synthetic experiments meant to validate the theory, post-training fine-tuning of standard models, and pre-training of Llama-style models of up to 800M parameters from scratch. Our results demonstrate that CAGE leads to state-of-the-art results across all scenarios: for fine-tuning QAT, it roughly halves the quantization error relative to the prior best method, whereas, in pre-training experiments, CAGE-trained models with 3-bits weights and activations (W3A3) offer lower loss than 4-bit (W4A4) models trained with the prior best method (QuEST~\citep{panferov2025quest}). 

\item Moreover, we show experimentally that the CAGE approach is \textit{optimizer-agnostic}, as it leads to consistent gains across several optimizers, including AdamW~\citep{loshchilov2018adamw}, Muon~\citep{muon} and Shampoo~\cite{shampoo}. 
\end{itemize}

In summary, we show that incorporating curvature-aware gradient correction via CAGE is a promising step toward bridging the remaining performance gap between low-bit quantized models and their full-precision counterparts.

%% file: related.tex
The central challenge in QAT is the non-differentiable quantization function, with zero gradients almost everywhere. The standard is the STE, initially suggested by~\citet{Hinton2012} and formalized by~\citet{bengio2013ste}, which bypasses the quantization operator during the backward pass. Yet, STE is known to often lead to instability and suboptimal convergence~\citep{Yin2019}.

Thus, substantial research has focused on STE refinements through learnable quantization parameters \citep{esser2020lsq}, improving the accuracy of the gradient itself \citep{Le2021AdaSTE} or incorporating loss regularization with proximal mapping \citep{bai2018proxquant}. For instance, AdaSTE~\citep{Le2021AdaSTE} enhances gradient estimation by viewing QAT problem from the perspective of bi-level optimization. ReSTE~\citep{Wu2023ReSTE} balances the estimation error and the gradient stability via a rectified estimator. \citet{liu2023bridging} proposed Reinmax, noticing that the STE acts as a first-order approximation of the gradient and proposed a second-order estimator via Heun's method.

In addition, ProxQuant~\citep{bai2018proxquant} reformulates QAT as a regularized learning problem solved via proximal gradient methods. Instead of relying solely on STE, ProxQuant applies a prox operator between stochastic gradient steps to encourage quantization. Similar to ProxQuant, PARQ uses convex and piecewise-affine regularization to induce weight quantization with the difference that gradient updates are added to the full-precision parameters before applying the proximal operator \cite{jin2025PARQ}. 

The Mirror Descent formulation interprets full-precision parameters as the dual of quantized ones and analyses the cases where quantization operator generates a valid mirror map. Particularly, if we take softmax projection as quantization operator, then the dual iterates of STE follow Exponentiated Gradient Descent (EGD) \cite{Ajanthan2021Ajanthan}. Markov Random Field (MRF) and lifted probability space formulations \cite{Ajanthan2019MRF} represent each parameter of the model as weighted average of quantized values, and minimizes the loss with respect to these probabilistic weights.

\citet{lee2021elementwisescaling} proposed the element-wise gradient scaling (EWGS) approach, in which the scaling of each gradient element depends on both its sign and the corresponding quantization error. Additionally, these scalings are adjusted by a global scaling factor obtained from approximate Hessian information. A related method employs position-based scaled gradients, where the gradient estimator is adjusted only by multiplying the quantization error with a small positive margin \cite{kim2020positionbasedscaling}.

Another phenomenon observed in STE-based training is the oscillation of weights around decision boundaries between adjacent quantization grid points \cite{défossez2022diffmodelcompression}. To mitigate these oscillations and stabilize training, \citet{nagel2022oscillationsQAT} introduced a regularization term on top of the loss to dampen the oscillations or alternatively proposed mechanisms for detecting oscillatory weights and freezing them during training.

By contrast to prior work, \methodname{} provides a multi-objective optimization perspective for QAT. Instead of refining the first-order approximation or applying standard regularization, \methodname{} introduces a curvature-aware correction term to the gradient.

%% file: method.tex
\subsection{Problem Description and Motivation}

Intuitively, the key question in quantization-aware training (QAT) is to construct a gradient estimator that updates the parameters towards an ``optimal'' quantized model, i.e., a constrained model which minimizes the loss. More precisely, the optimization problem underlying QAT can be described by the following constrained optimization:
\begin{equation}\label{main-problem-1}
\min_{x\in\mathcal{Q}} f(x),    
\end{equation}

where $\mathcal{Q} = \{x\in\R^d \colon x = Q(x)\}$ is the constraint set of quantized parameters and $Q\colon\R^d\to\R^d$ is the quantization operator. An alternative way of looking at the same problem is to consider its unconstrained reformulation and minimizing the objective $f(Q(x))$ with respect to $x\in\R^d$ \citep{Li2017BinaryConnect}. 
The key issue here is the non-differentiability of quantization operator $Q$. 
As such, the straight-through estimator of \citep{bengio2013estimating} is the default choice to approximate the gradient of $f(Q(x)$. 
Specifically, $\nabla_x [f(Q(x))] = JQ(x)^\top \cdot \nabla f(Q(x))$ is approximated by $\nabla f(Q(x))$. Effectively, this replaces the Jacobian of the quantization operation by the identity matrix~\cite{Li2017BinaryConnect,Hou2019QM,Chen2021QAdamEF}.

{\bf Error feedback reformulation.} One way to gain more insight into the dynamics of STE is to examine it through the lens of error feedback. It is easy to observe that SGD-based training with STE over the iterate $x_t$ is equivalent to an instance of error-feedback~\cite{chen2021quantizedadamerrorfeedback} with quantized parameters $w_t = Q(x_t)$ and quantization error $e_t = x_t - Q(x_t)$. This equivalence is described over iterations $t \geq 0$ below:
\begin{equation*}
    x_{t+1} = x_t - \alpha \widetilde{\nabla} f(Q(x_t))
    \iff
    \begin{cases}
        g_t = \alpha\widetilde{\nabla}f(w_t) - e_t 

        \\
        w_{t+1} = Q(w_t - g_t) \\
        e_{t+1} = (w_t - g_t) - w_{t+1}.
    \end{cases}
\end{equation*}

Effectively, the right-hand-side formulation says that STE is equivalent to a process where the quantization error $e_{t + 1}$ due to weight quantization at a given step $(w_t - g_t) - w_{t+1}$ is fed into the \emph{gradients} at the next step. This feedback loop would be ``exact'', fully correcting the error, if the loss function $f$ were an isotropic quadratic function, i.e. with identity Hessian, but is not the right update in general. 

{\bf Multi-objective perspective.} Prior work on convergence guarantees for the problem described in Equation~\eqref{main-problem-1} in the non-convex setting attempts to find a quantized point $Q(x^*)$ that is a stationary point for the loss $f$, namely $\nabla f(Q(x^*)) = 0$. 

However, as quantization is not an invertible transformation, this goal cannot be achieved in general, and current convergence guarantees have non-vanishing terms in their rate, proportional to quantization error \cite{lin2020dynamicmodelpruningfeedback,chen2021quantizedadamerrorfeedback,panferov2025unifiedscalinglawscompressed}. 

For the sake of illustration, consider a toy example where our scalar loss function is $f(x) = \frac{1}{2}(x - \frac{1}{2})^2$, and the quantization operator $Q(x) = \floor{x}$ gives the integer part of the one dimensional input $x\in\R$. Clearly, $\nabla f(Q(x)) = \floor{x}-\frac{1}{2}$ does not vanish at any point, with the smallest absolute value being $\frac{1}{2}$, coming from the quantization/rounding error.

Instead, we propose to view the QAT problem in Equation~\eqref{main-problem-1} from the perspective of multi-objective optimization. The first objective is to minimize the loss $f(x)$: in the non-convex setting this becomes finding a stationary point, namely solving $\nabla f(x)=0$. The second objective is to satisfy the quantization constraint $x\in{\mathcal Q}$. In other words, we aim to find a solution $x^*$ for which $\nabla f(x^*) = 0$ and the distance between $x^*$ and $Q(x^*)$ is minimized. As noted before, it is easy to see that these two conditions do not have to hold in general, meaning $\nabla f(x)$ can be non-zero for any quantized point $x\in{\mathcal Q}$. Instead, we define solutions as {\em Pareto optimal} points that balance these two conditions.

Let $x\in\R^d$ be the current state of parameters. If we update towards $-\nabla f(x)$ (or any other direction that aligns with it) with small enough step-size, then we would minimize the loss. Clearly, the opposite direction would increase the loss. Analogously, if we update towards $Q(x)-x$, then we would reduce quantization error. To have those two objectives in balance, we want our iterates to converge to some Pareto-optimal state $x^*$ such that $\nabla f(x^*) = \lambda(Q(x^*) - x^*)$ for some scalar $\lambda > 0$. In this case, any sufficiently small update applied to $x^*$ would hurt at least one of the objectives. In other words, any local improvement of one objective would be at the cost of other. Therefore, we define {\em $\lambda$-Pareto optimal} solution $x^*$ for the problem \eqref{main-problem-1} as
\begin{equation}\label{pareto-opt}
    \nabla_{\rm \lambda P} f(Q(x^*)) \eqdef \nabla f(x^*) + \lambda(x^* - Q(x^*)) = 0,
\end{equation}
where $\lambda>0$ is a parameter balancing the two objectives. Recalling the toy example mentioned earlier, the value $x^*(\lambda) = \frac{1}{2(1+\lambda)}$ is $\lambda$-Pareto optimal for any $\lambda>0$. Note that there can be many Pareto optimal solutions for the same $\lambda$. In our toy example, $x^*=-\nicefrac{1}{4}$ is another $1$-Pareto optimal value.

Motivated by this multi-objective perspective and by the new optimality condition in Equation~\eqref{pareto-opt}, we propose incorporating the quantization error $e_t = x_t - Q(x_t)$ directly into the training dynamics. By doing so, we implicitly regularize the initial task loss, which can be made explicit if the quantization scheme is smooth (as we demonstrate in the theoretical analysis). Indeed, under the smoothness assumption of $Q$ (Assumption 3), the quantization error can be expressed as $x - Q(x) = \nabla \phi(x)$ for some smooth regularizer $\phi(x)$.
Notably, our approach naturally avoids the need to handle the (non-existent) Jacobian $JQ(x)$ of the quantization operator and bypasses the computation of a proximal step for the explicit ``quadratic'' regularizer $\frac{\lambda}{2}\|x - Q(x)\|_2^2$, which would be infeasible for the dynamic quantization used in our experiments \cite{bai2018proxquant}.

\paragraph{Error coupling.} Noting that our approach is optimizer-agnostic, we propose two ways to incorporate the quantization error, referred to as the {\em coupled} and {\em decoupled} corrections. The distinction lies in whether the quantization error is added to the current gradient before it is passed to the optimizer (coupled correction) or added on top of the model update computed by the optimizer (decoupled correction).

If the base optimizer is SGD, then these two corrections are identical. We will provide theoretical convergence guarantees to a Pareto-optimal point in the smooth non-convex setting. However, for stateful optimizers such as Adam, these two corrections produce conceptually different schemes, which we discuss below (subsuming bias correction terms into the learning rate $\alpha$):
\begin{align*}
& \textnormal{Quantization error and mini-batch gradient} \\
& \qquad e_t = x_t - Q(x_t) \\ 
& \qquad g_t = \widetilde{\nabla}f(x_t) {\color{ForestGreen} \;+\; \lambda_t e_t},\; {\color{ForestGreen} \textrm{(coupled correction)}} \\
& \textnormal{Optimizer states} \\
& \qquad m_t = \beta_1 m_{t-1} + (1-\beta_1)g_t, \\
& \qquad v_t = \beta_2 v_{t-1} + (1-\beta_2)g_t^2, \\
& \textnormal{Optimizer update} \\
& \qquad \Delta_t = m_t/(\sqrt{v_t} + \varepsilon) {\color{blue} \;+\; \lambda_te_t},\; {\color{blue}\textrm{(decoupled correction)}} \\
&\textnormal{Model update} \\
& \qquad x_{t+1} = x_t - \alpha \Delta_t,
\end{align*}

To compare these two correction variants in the context of the Adam optimizer, note that the coupled correction term ${\color{ForestGreen} \lambda_t e_t}$ becomes $(1 - \beta_1){\color{ForestGreen} \lambda_t e_t} / (\sqrt{v_t} + \varepsilon)$ after being processed by the optimizer states. In contrast to the decoupled correction, the coupled version introduces diagonally-preconditioned curvature-aware correction term, where the curvature is approximated using the Adam statistics already maintained within the optimizer states.

Crucially, both variants apply the same Pareto-derived correction \emph{direction} $e_t = x_t - Q(x_t)$. They differ only in how the correction is scaled.
In the coupled variant, the correction term ${\color{ForestGreen} \lambda_t e_t}$ is absorbed into the gradient before preconditioning, so it effectively becomes $(1 - \beta_1){\color{ForestGreen} \lambda_t e_t} / (\sqrt{v_t} + \varepsilon)$ after being processed by the optimizer states. This introduces a diagonally-preconditioned, \emph{curvature-aware} correction, where the curvature is approximated using the Adam second-moment statistics already maintained within the optimizer.
The decoupled variant, by contrast, applies the correction at unit scale after the optimizer step, deliberately avoiding the preconditioning path. This is a practical stability choice as it prevents the correction magnitude from being distorted by per-coordinate variance estimates (which can fluctuate in low-bit regimes) while preserving the same Pareto-stationarity direction. We show in Appendix~\ref{app:de_vs_coupled_cage} that both variants achieve nearly identical final perplexity across model sizes and quantization configurations under AdamW, confirming that the primary gains originate from the shared correction direction rather than the specific scaling.

\paragraph{Relationship to LOTION.} 
Concurrent work~\cite{kwun2025lotion} proposes an approach called LOTION, which smooths the quantized loss using unbiased randomized rounding ({\rm RR}) via $f(Q(x)) \approx \E_{\epsilon \sim {\rm RR}}[f(x + \epsilon)]$, and considers its second-order expansion as a surrogate loss to minimize, i.e., $f(x + \epsilon) \approx f(x) + \epsilon^{\top}\nabla f(x) + \tfrac{1}{2}\epsilon^{\top}{\bf H}(x)\epsilon$. Due to the unbiasedness of rounding (i.e., $\E[\epsilon] = 0$), the first-order term of the expansion vanishes in expectation, and the smoothed loss becomes the original loss $f(x)$ with an additional curvature-aware regularization term $\tfrac{1}{2}{\rm Tr}\left({\bf H}(x)\,{\rm Cov}[\epsilon]\right)$. Here, the exact Hessian matrix is replaced by its Gauss–Newton approximation. If we further apply the empirical Fisher approximation and approximate the diagonal entries of the Gauss–Newton matrix using Adam statistics, this regularization takes similar form as our coupled correction variant (except the square root). Exactly computing the gradient estimator for the regularized loss would require differentiating the regularizer, which would involve third-order derivatives of the loss and is impractical. As such, this step is proposed to be skipped by the authors. 

In contrast to LOTION, we propose to regularize the training dynamics directly, rather than the loss itself. One advantage of this approach is that the regularized dynamics naturally arise from our Pareto-optimality condition, while the practical benefit is that it avoids the need to differentiate or compute the proximal operator of a loss regularizer. In addition, as described in their paper, LOTION requires a full-precision forward pass, and the final model is a weight-only quantized model. By contrast, \methodname{} training is done with weights and activations quantized, where it can also benefit from high-performance low-precision GEMM kernels for more efficient training and inference.

\subsection{Theoretical Analysis}

In this section, we present our main convergence result for \methodname{} when the base optimizer is SGD. In this case, coupled and decoupled versions coincide with the following iterates:
\begin{equation}\label{iter-1}
    x_{t+1} = x_t - \alpha (\widetilde{\nabla} f(x_t) + \lambda(x_t - Q(x_t)))
\end{equation}
for $t=0,1,\dots$. We make the following assumptions regarding the smoothness of the loss and the nature of the stochastic noise.

\begin{assumption}\label{asm:smooth} 
    The loss function $f\colon\R^d\to\R$ is lower bounded by some $f^* \in \mathbb{R}$ and is $L_f$-smooth, namely,
    $\|\nabla f(x) - \nabla f (y)\|_2 \leq L_f \|x-y\|_2,\; \text{for any } x,y\in\R^d.$ 
\end{assumption}
\begin{assumption}\label{asm:unbiased}
    For all iterates $t$, the stochastic gradient $\widetilde{\nabla}f(x_t)$ is unbiased, namely $\E[\widetilde{\nabla}f(x_t)] = \nabla f(x_t)$, and the variance is bounded $\E[\|\widetilde{\nabla}f(x_t) - \nabla f(x_t)\|_2^2] \le \sigma^2$ for some constant $\sigma^2\ge0$.
\end{assumption}
Both assumptions are quite standard in the optimization literature and, in some sense, minimal to get meaningful convergence guarantees. Additionally, to handle the quantization operator $Q$ in the analysis, it is common to approximate the actual non-smooth quantization operator $Q$ with smooth surrogates with annealing hyperparameters.

\begin{assumption}\label{asm:quant-field}
    There exists some  $L_{\phi}$-smooth function $\phi\colon\R^{d}\to\R$ such that the quantization error $x-Q(x) = \nabla\phi(x)$.
\end{assumption}

Note that Assumption \ref{asm:quant-field} is essentially equivalent to the Lipschitz continuity of the quantization operator $Q$. The purpose of formulating this condition using an auxiliary function $\phi(x)$ is to hint that the iterates in \eqref{iter-1} aim to minimize the regularized loss $f(x) + \lambda\phi(x)$, as we elaborate on later in the analysis. Next, we present the convergence result.

\begin{theorem}\label{thm:main-pareto}
    Let Assumptions \ref{asm:smooth}, \ref{asm:unbiased}, \ref{asm:quant-field} hold and $L = L_f + \lambda L_{\phi}$. Then, for any $\lambda\ge0$, the iterates \eqref{iter-1} with step-size $\alpha = \min(\frac{1}{L}, \frac{1}{\sqrt{T}})$ satisfy

    $$
    \E \|\nabla_{\rm \lambda P} f(Q(\hat x))\|^2
    \le \frac{2}{\sqrt{T}} \left( A_1 + A_2 + A_3 \right) \max\left(1, \tfrac{L}{\sqrt{T}}\right),
    $$
    where $\hat x$ is a random iterate drawn from the history $\{x_0,x_1,\dots,x_{T-1}\}$ uniformly at random and 
    $$
    A_1 = f(x_0) - f^*,\; A_2 = \frac{L \sigma^2}{2}
    $$
    $$
    A_3 = \lambda\max_{x\in[x_0,x_T]} \|x-Q(x)\| \|x_0-x_T\|.
    $$
\end{theorem}

\paragraph{Discussion.} First, observe that the convergence bound shows strong ergodic convergence to a Pareto-optimal state of the problem in Equation~\ref{main-problem-1}, with a convergence rate that is optimal in the unquantized case. The convergence measure is defined with respect to our proposed Pareto-gradients (Equation~\eqref{pareto-opt}) and, importantly, there is no non-vanishing term in the upper bound because of quantization error. In practical settings, both the distance $\|x_0 - x_T\|$ and the quantization error $\|x - Q(x)\|$ are expected to remain bounded for any total number of training steps $T$. Therefore, this implies that the iterates in Equation~\eqref{iter-1} achieve $\mathcal{O}(\frac{1}{\sqrt{T}})$ ergodic convergence to the Pareto-optimal solution.

\subsection{Practical Implementation}
\label{sec:impl-cage}

Since our main application is in training and fine-tuning LLMs, we implement \methodname{} on top of AdamW~\cite{loshchilov2018adamw}. In contrast to LOTION~\cite{kwun2025lotion}, we focus on the \emph{decoupled} update rule and keep the correction term outside the preconditioning path (Alg.~\ref{alg:cage-adamw}). We use this choice in most experiments, as we have found it to work better in practice. The base optimizer AdamW processes the mini-batch gradients $g_t$ obtained from STE, and the \methodname{} correction acts as a lightweight, elementwise post-step on the parameters. This decoupling has multiple practical advantages: (1) it preserves the optimizer's well-understood behavior, (2) it keeps the Pareto stationarity direction $x-Q(x)$ from being distorted by preconditioning, which empirically stabilizes training in low-bit regimes, and (3) it avoids numerical errors by decoupling terms with different behavior.

\begin{algorithm}[t]
\caption{\methodname{}-AdamW (decoupled)}
\label{alg:cage-adamw}
\begin{algorithmic}[1]
\Require Initial parameters $x_0$; total steps $T$; AdamW hyperparameters $\beta_1,\beta_2,\alpha,\omega,\varepsilon$; Quantization function $Q$;
\Require CAGE coefficient $\lambda$, silence ratio $s$
\Ensure Parameters $x_T$
\State Initialize $m_0 \gets 0,\; v_0 \gets 0,\; e_0 \gets 0$
\For{$t=1,2,\dots,T$}

  \State $r_t \gets t/T$ \Comment{training progress ratio}
  \If {$r_t \le s$}
    \State \textcolor{blue}{$\lambda_t \gets 0$}
  \EndIf  
    \If {$r_t > s$}
  \State \textcolor{blue}{$\lambda_t \gets \lambda \cdot 
  \dfrac{r_t - s}{1-s}$}
  \EndIf

  \State \emph{Sample minibatch and compute stochastic gradient $g_t$ with quantized forward pass.}
  \State $x_t \gets (1-\alpha \omega)\,x_t$ \Comment{decoupled weight decay}
  \State $m_t \gets \beta_1 m_{t-1} + (1-\beta_1) g_t$
  \State $v_t \gets \beta_2 v_{t-1} + (1-\beta_2)\, g_t \odot g_t$
  \State $\hat m_t \gets m_t/(1-\beta_1^t)$;\quad $\hat v_t \gets v_t/(1-\beta_2^t)$
  \State $\tilde x_{t+1} \gets x_t - \alpha\, \hat m_t / \left(\sqrt{\hat v_t}+\varepsilon\right)$
  \State \textcolor{blue}{$e_t \gets x_t - Q(x_t)$ \Comment{quantization error (no grad)}}
  \State \textcolor{blue}{$x_{t+1} \gets \tilde x_{t+1} - \alpha \,\lambda_t\, e_t$ \Comment{decoupled correction}}
\EndFor
\State \Return $x_T$
\end{algorithmic}
\end{algorithm}

\paragraph{Decoupled update.}
Given current set of parameters  $x_t$ and STE gradient estimator $g_t$ through backpropagation, AdamW performs
$$
\tilde x_{t+1} = x_t -\alpha\cdot\frac{\hat m_t}{\sqrt{\hat v_t}+\varepsilon},
\text{ with }
\hat m_t=\frac{m_t}{1-\beta_1^t},\;
\hat v_t=\frac{v_t}{1-\beta_2^t}
$$
with additional decoupled weight decay regularization in the form of $x_t \leftarrow (1-\alpha\omega)x_t$.\footnote{We use the common formulation where weight decay is applied before the Adam step; other equivalent placements are fine so long as decay is not mixed into $g_t$.}
The \methodname{} correction term then pushes the parameters toward the quantized support via the instantaneous quantization error:
$$
e_t \coloneq x_t - Q(x_t),\qquad
x_{t+1} = \tilde x_{t+1} - \alpha\lambda_t\bar e_t.
$$

\paragraph{Warmup schedule.}
In practice, we have found that activating the correction from the very beginning of training can over-constrain the dynamics before the model settles into the loss landscape. Similar to standard learning-rate warmup schedules, we introduce a \emph{silence period} ratio $s \in [0,1)$, which skips the correction for the initial $sT$ steps and then linearly ramps it up to its target magnitude:
$$
r_t \coloneq \frac{t}{T},\qquad
\lambda_t =
\begin{cases}
0, & r_t \le s,\\
\lambda \cdot \dfrac{r_t - s}{1-s}, & r_t > s.
\end{cases}
$$

The rationale for the silence period is as follows. Early in training, the iterate traverses many quantization cells in rapid succession. The quantization residuals $e_t$ behave like approximately mean-zero, bounded perturbations that cancel in aggregate (analogous to a random walk across cell boundaries). Applying the correction in this regime is therefore ineffective as the cumulative correction has $O(\sqrt{T})$ fluctuations but near-zero mean, and can destabilize training by injecting noise-like forcing into the dynamics. In contrast, near convergence, the iterate settles into a small neighborhood and the quantized anchor $Q(x_t)$ remains stable for many consecutive steps. In this regime the residuals become \emph{coherent} in the sense that they accumulate rather than cancel, and the correction effectively pushes parameters toward the quantized support, which is precisely when enforcing Pareto-stationarity is beneficial. We provide a formal analysis of this early-cancellation vs late-coherence argument, including concentration bounds, in Appendix~\ref{app:lambda_schedule}.

\begin{figure*}[ht!]
\centering

\begin{minipage}[ht]{0.47\textwidth} 
    \includegraphics[width=0.9\columnwidth]{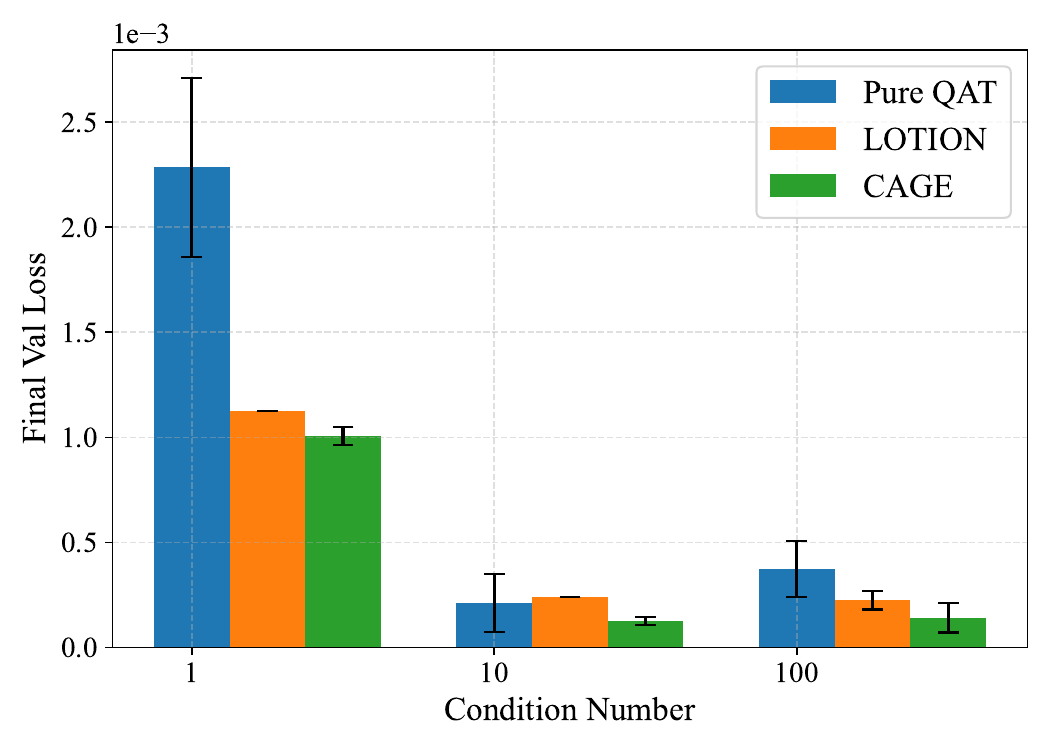}
    \caption{Quadratic with non-isotropic Hessian: final validation loss for all methods various condition numbers (lower is better). The loss for \methodname{} does not include the regularizer term. We find that \methodname{} consistently reduces stationary error.}
    \label{fig:quadratic_training_bars}
\end{minipage}%
\hfill
\begin{minipage}[ht]{0.47\textwidth} 
    \centering
    \includegraphics[width=0.8\linewidth]{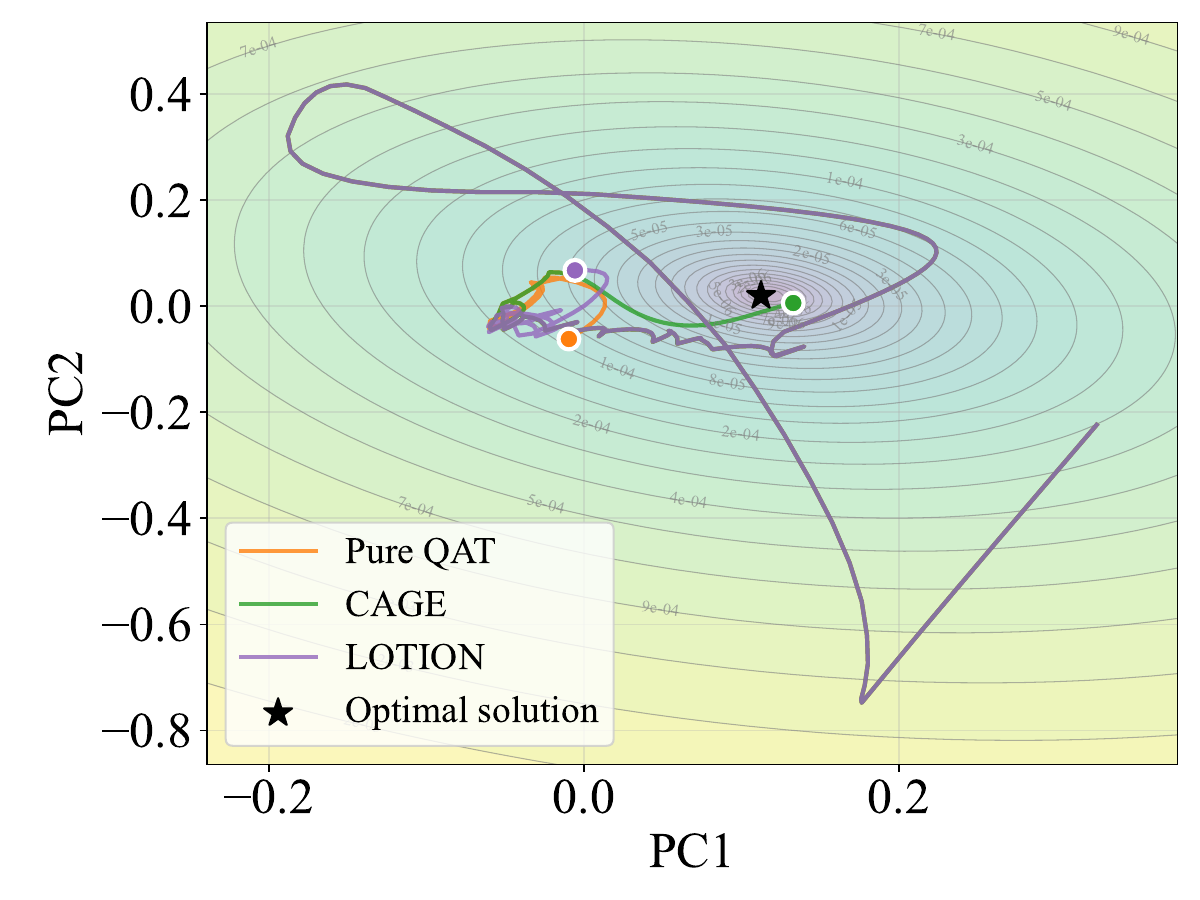}
    \caption{Representative trajectories (projected to the top two principal components PC1 and PC2) for the quadratic task; \methodname{} follows curvature and converges closest to the optimum.}
    \label{fig:quadratic_training_trajectory}
\end{minipage}
\end{figure*}

In practice, we have found the parameters $s\in[0.8,0.95]$ with a linear ramp and $\lambda \in [1, 10]$ (depending on the model) to work robustly for AdamW. 
We provide sweeps justifying these choices in Appendix~\ref{app:cage_ablation}.
The ramp prevents abrupt shifts in the optimization objective and reduces loss spikes in training.
The hyperparameter sweep in Appendix~\ref{app:cage_ablation} confirms that CAGE is insensitive to the exact choice of $s$ and $\lambda$ within this range, i.e., these are coarse knobs rather than sensitive tuning parameters, with most of the configurations in the recommended region consistently improving over the baseline.

%% file: experiments.tex
\paragraph{Quantization pipeline.}
Unless otherwise stated, our experiments use a standard \emph{row-wise} matrix quantizer, e.g.~\cite{qlora}. Specifically, our QAT instantiation is based on QuEST~\cite{panferov2025quest}. This works as follows: let \(x\in\mathbb{R}^{d}\) denote a tensor row, and let \(H\in\{\pm \tfrac{1}{\sqrt{d}}\}^{d\times d}\) be a Hadamard transform of corresponding size. Before quantization,  each tensor is ``rotated'' via the orthogonal Hadamard matrix: 
\[
z = Hx.
\]
Given the target bit-width \(b\), we quantize symmetrically with integer bounds \(q_{\max}=2^{b-1}-1\) and \(q_{\min}=-2^{b-1}\). Using the pre-computed MSE-optimal Gaussian clipping factor \(k_b\) (per bit-width), with \(\sigma=\sqrt{\tfrac{1}{d}\sum_i z_i^2}\), we set the scale
\[
s = \frac{k_b\,\sigma}{q_{\max}}.
\]
We then apply the quantization operation, rescale, and invert the transform: 
\[
q = \operatorname{clip}\left(\left\lfloor \tfrac{z}{s}\right\rceil, q_{\min}, q_{\max}\right),\qquad \hat z = sq,\qquad \hat x = H^\top \hat z,
\]
where \(\lfloor\cdot\rceil\) denotes round-to-nearest. The backward pass uses QuEST's trust-masked STE in the transform domain.

Note that \methodname{} itself is \emph{quantizer-agnostic}. It augments the optimizer update via the instantaneous quantization error \(e_t=x_t-Q(x_t)\) (see §\ref{sec:impl-cage}). We choose the QuEST quantizer here because it is a strong SOTA baseline for low-bit training. 
We note that the QuEST paper already showed superior results in terms of both training stability and loss, relative to prior methods such as LSQ/LSQ+~\cite{esser2020lsq, bhalgat2020lsqplus}. As such, we omit comparisons with further prior work, while noting that other quantizers can be used with \methodname{} without changing the method.

\paragraph{Low-bit system support.}
Our training implementation uses \emph{simulated quantization} (quantize $\to$ dequantize $\to$ standard-precision matmuls), which is the standard practice in QAT research~\cite{panferov2025quest,kwun2025lotion}. No custom low-bit GEMM kernels are used during training.

For \textbf{W4A4}, real kernel-level support already exists on mainstream GPU hardware. NVIDIA's Blackwell architecture provides native FP4 acceleration via 5th-generation Tensor Cores, supporting both MXFP4~\cite{mxspec} and NVFP4~\cite{nvidia2025nvfp4}. The fine-tuning experiments in \S\ref{sec:experiments} specifically target the MXFP4 format for this reason, though CAGE is equally applicable to NVFP4-based pipelines.
For \textbf{W3A3} and \textbf{W2A2}, no widely-available hardware support existed at the time of writing. We include these precisions because QAT research should push algorithmic frontiers across the full precision spectrum, and emerging hardware is beginning to target these regimes. Our result showing W3A3 CAGE matching W4A4 QuEST suggests that better QAT methods can make sub-4-bit deployment practically viable as hardware support matures.

We emphasize that the main contribution of \methodname{} is \emph{not} a new kernel. It is a near-cost-free correction term added to the optimizer step that improves accuracy for \emph{any} underlying quantizer. \methodname{} requires only access to the quantization residual $e_t = x_t - Q(x_t)$, regardless of how $Q$ is implemented or at what precision. The various bitwidths and quantization formats in our experiments serve to demonstrate this generality.

\subsection{Synthetic Loss Experiments}

To examine the exact QAT dynamics, we first study different methods optimizing a quadratic objective \(f(x)=\tfrac{1}{2}x^\top A x - b^\top x\) with \(\kappa(A)\in\{1,10,100\}\) (non-isotropic curvature) over a quantized domain. 
We compare STE-SGD, STE-Adam, and CAGE-Adam (decoupled) under a 4-bit quantization of \(x\), using the same pipeline as above, under different condition numbers. 

More precisely, we initialize \(x_0\sim\mathcal{N}(0,\sigma_0^2 I)\), run a fixed budget of \(T\) steps, and measure (i) final validation loss over 10 seeds and (ii) trajectories in the \((x^\top A x)^{1/2}\) vs. iteration plane relative to the true minimizer \(x^\star=A^{-1}b\). Figure~\ref{fig:quadratic_training_bars} reports the distribution of final losses (mean\(\pm\)std across 10 seeds).  Figure~\ref{fig:quadratic_training_trajectory} shows representative trajectories and end-points over a reduced-dimension loss landscape around the optimum (following the top two principal components (PC)).
Figure~\ref{fig:quadratic_training_bars} shows that, across all condition numbers, CAGE reduces error in a statistically-significant way, while Figure~\ref{fig:quadratic_training_trajectory} shows that CAGE finds a quantized solution that is strictly closer to the optimum.

\begin{figure*}[t]
\centering

\begin{minipage}[ht]{0.47\textwidth} 
\includegraphics[width=0.95\columnwidth]{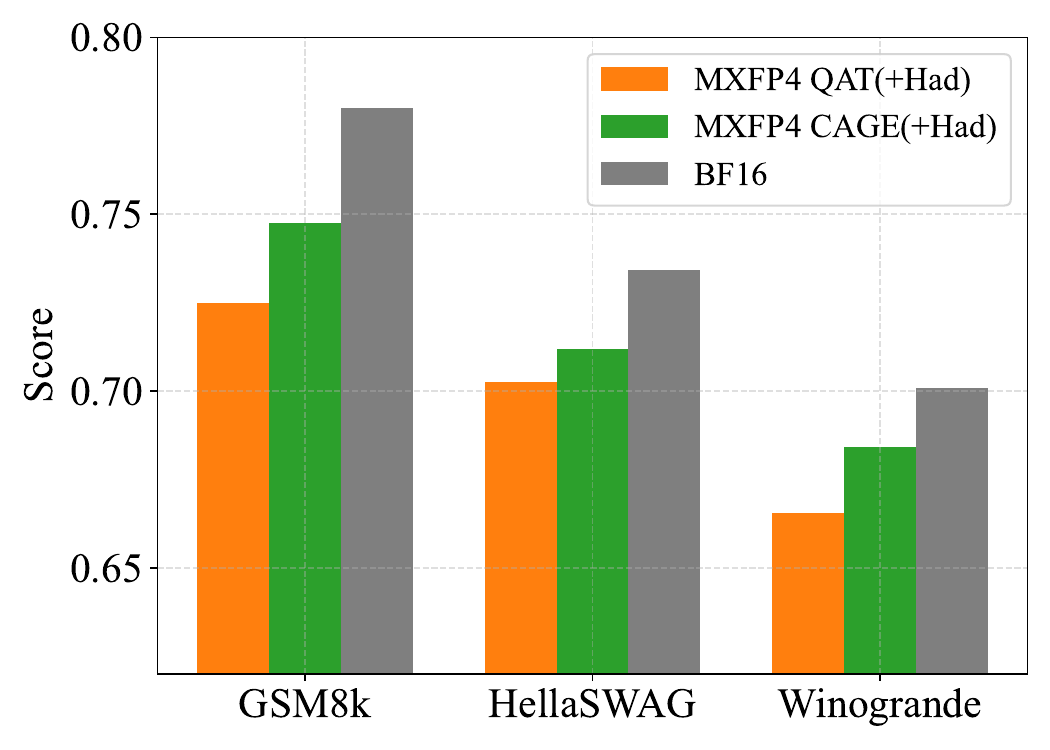}
\caption{QAT accuracy on Llama-3.2-3B (Tulu-SFT) for CAGE vs. the state-of-the-art MXFP4 baseline, using QuEST~\cite{panferov2025quest}. CAGE consistently improves GSM8K/HellaSwag/WinoGrande results, essentially halving the error due to quantization. Hadamard transform is applied in both cases for outlier mitigation.}
\label{fig:mxfp4_bench}
\end{minipage}%
\hfill
\begin{minipage}[ht]{0.47\textwidth} 
    \centering
    \includegraphics[width=\columnwidth]{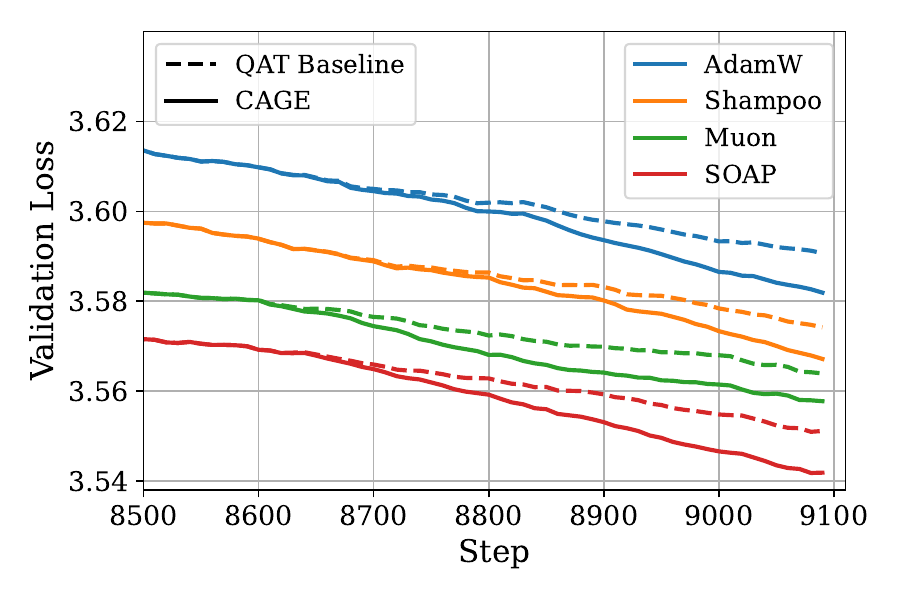}
    \vspace{-1em}
    \caption{Validation loss during QAT (W4A4) for AdamW, Shampoo, SOAP, and Muon, for standard QAT and CAGE. Solid lines denote training with CAGE, dashed -- without. Across all optimizers, the method yields lower validation loss.}
    \label{fig:optimizers-loss-curves}
\end{minipage}
\end{figure*}

\subsection{Quantization-Aware Fine-tuning}

Next, we evaluate the suitability of \methodname{} in a more realistic setting, where it is used to perform QAT over the MXFP4 4-bit floating format \cite{mxspec} on the Llama-3.2-3B model~\cite{llama3}. 
We fine-tune on Tulu-SFT \cite{lambert2024tulu3} with QAT (master weights in FP32, the forward pass has both weights and activations quantized to MXFP4), following the recently-proposed QAT recipe of~\cite{egiazarian2025bridginggappromiseperformance}, which yields state-of-the-art results for MXFP4. 

We report zero-shot or few-shot scores on GSM8K (exact-match), HellaSwag (accuracy), and WinoGrande (accuracy). (Higher is better.) 
Hyperparameters for the experiments are in Appendix \ref{app:hyperparameters}, where we also include the results without the Hadamard transform, which show similar trends.

Figure~\ref{fig:mxfp4_bench} summarizes the results. We observe that CAGE consistently improves accuracy relative to the state-of-the-art QAT baseline; specifically, it roughly \textit{halves} the quantization error when mapping the model to MXFP4 in this format.

\begin{figure*}[t]
\centering

\begin{minipage}[ht]{0.6\textwidth}
    \centering
    \includegraphics[width=\linewidth]{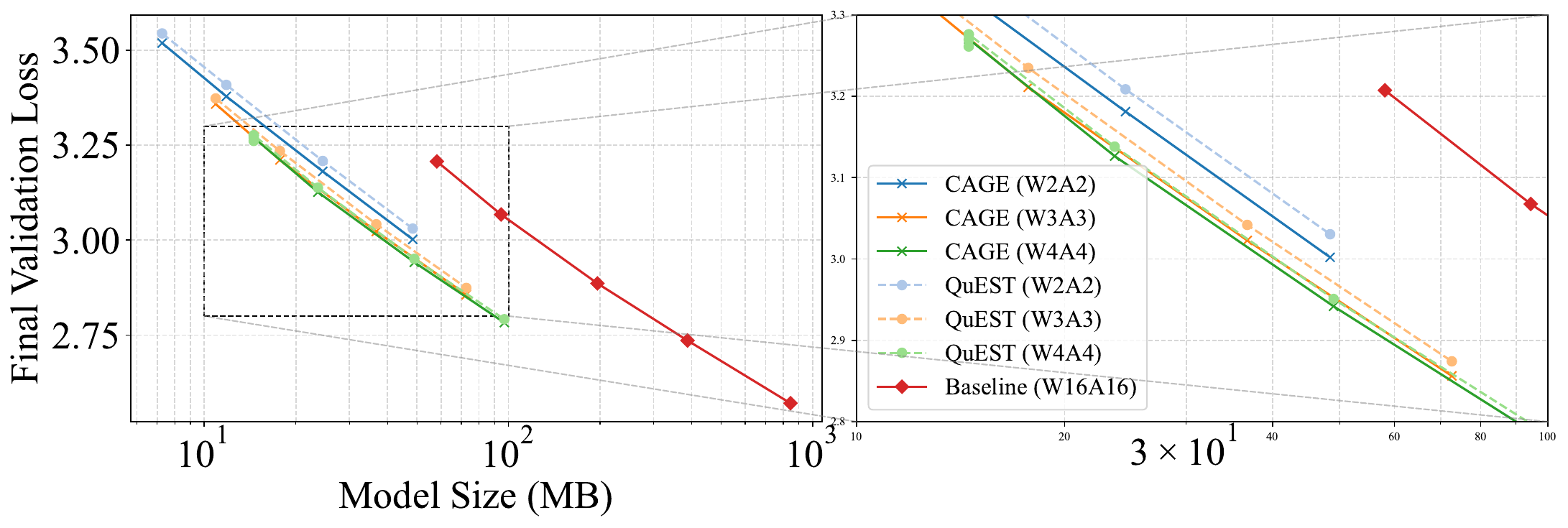}
    \caption{Validation loss versus model size (bytes) for weight and activation quantization using W2A2, W3A3 and W4A4, comparing CAGE+QuEST to baseline QuEST and BF16. Observe that CAGE yields  consistently lower loss.}
    \label{fig:scaling_law_fig}
\end{minipage}%
\qquad
\begin{minipage}[ht]{0.3\textwidth}

    \centering
    \includegraphics[width=\linewidth]{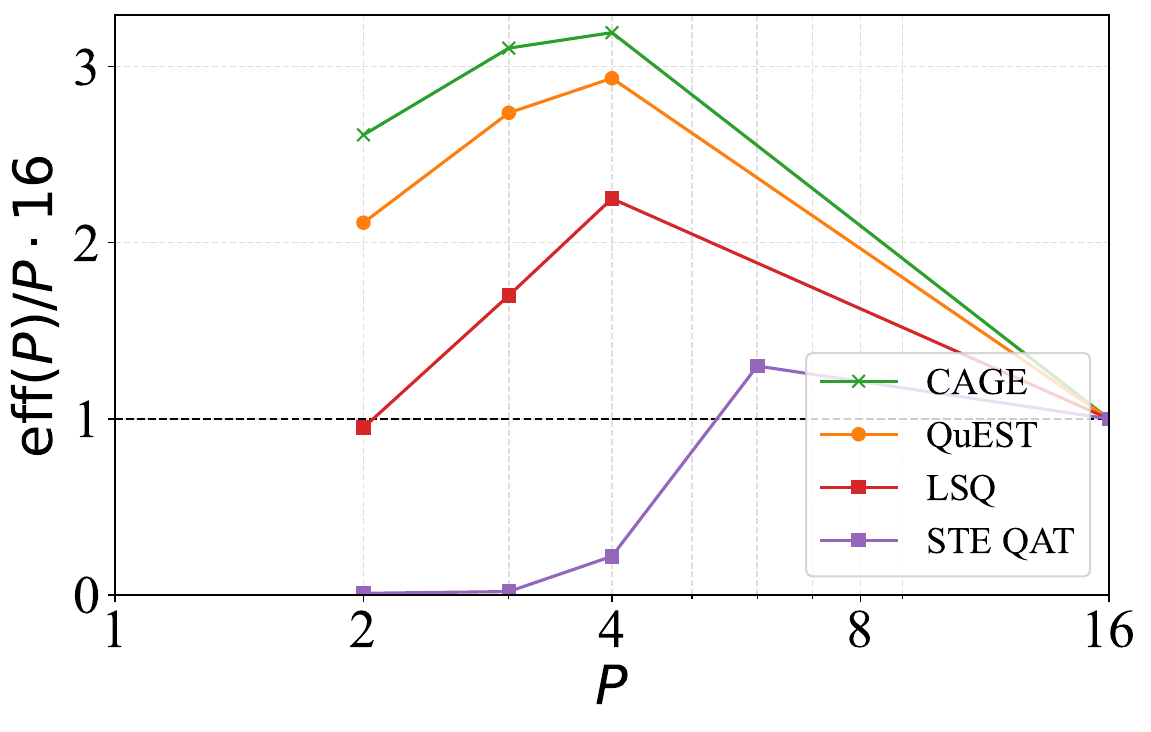}

    \caption{Fitted parameter efficiency $\mathrm{eff}(P)$, normalized relative to standard 16-bit, across precisions. CAGE improves effective  model capacity over QuEST, peaking near 4-bit.}
    \label{fig:eff_comparison}
\end{minipage}

\end{figure*}

\begin{table*}[h!]
\centering
\caption{Pretraining results (final validation perplexity; lower is better) for W4A4 across model sizes. HT = Hadamard transform. BF16 is a full-precision reference.}
\vspace{0.5em}
\label{tab:pretrain_w4a4}
\begin{tabular}{lcccccc|c}
\toprule
Method & 30M & 50M & 100M & 200M & 430M & 800M & \(\mathrm{eff}(P)\) \\
\midrule
CAGE + HT        & \textbf{26.277} & \textbf{22.747} & \textbf{18.944} & \textbf{16.166}   &  \textbf{13.789}   & \textbf{12.182} & \textbf{0.797} \\
QuEST + HT       & 26.475 & 23.062 & 19.123 & 16.311 &  14.169 & 12.482 & 0.733 \\
\midrule
CAGE (no HT)     & 27.287 & 23.781 & 19.630 & 16.596 &  14.024  & 12.341 & \textbf{0.705} \\
QuEST (no HT)    & 27.401 & 23.991 & 19.799 & 17.093 &  14.375 & 12.797 & 0.620 \\
\midrule
BF16 (reference) & 24.715 & 21.491 & 17.923 & 15.422 &  13.176   & 11.698 & 1.0 \\
\bottomrule
\end{tabular}
\end{table*}

\subsection{Pretraining Experiments} 

Next, we test our method by pre-training Llama-style transformers with parameter counts \(N\in\{30\text{M},50\text{M},100\text{M},200\text{M},430\text{M},800\text{M}\}\) under quantization-aware training from scratch, with weight/activation bit-widths \(b\in\{2,3,4\}\). 
Our baseline is again the QuEST QAT approach, which uses a Hadamard transform prior to quantization~\cite{panferov2025quest}, which was shown to be superior to classic methods such as STE and LSQ~\cite{esser2020lsq}. Training uses the C4 dataset~\citep{c4} with a fixed budget of \(D=100\times N\) tokens, for a token-per-parameter (TPP) ratio of 100, where \(D\) is the number of tokens and \(N\) is the number of model parameters. Master weights are stored FP32, and updates follow AdamW with decoupled weight decay \cite{loshchilov2018adamw} as in Algorithm~\ref{alg:cage-adamw}. 

 We compare \methodname{} against (i) QuEST  (the current SOTA baseline) and (ii) a high-precision BF16 training reference. Figure~\ref{fig:scaling_law_fig} shows validation loss vs. model size for \(b\in\{2,3,4\}\), with \methodname{}, QuEST baseline, and BF16.

Experiments are done on 8xH100 GPUs using standard hyperparameters (LR schedule, warmup, clipping, etc.), presented in Appendix~\ref{app:hyperparameters}. Each pretraining result is an average over three seeds and the mean and standard deviation are reported in Table \ref{tab:pretrain_w4a4}. Models up to $200\text{M}$ are visualized in Figure \ref{fig:scaling_law_fig}. We note that the standard deviation is too small to be visible in Figure~\ref{fig:scaling_law_fig}.

\paragraph{Loss comparison.} 
Table~\ref{tab:pretrain_w4a4} presents the actual validation loss results for W4A4 training across methods and model sizes. We observe that CAGE consistently improves upon QuEST, by significant margins, across all model sizes, and both with and without the Hadamard transform. 

Figure~\ref{fig:scaling_law_fig} examines the validation loss-vs-model-size Pareto front induced by CAGE and QuEST across W4A4, W3A3, and W2A2 precisions. 
We observe that CAGE is consistently Pareto-superior relative to QuEST, at every precision. 
W4A4 CAGE appears to be Pareto-optimal in terms of obtaining the lowest loss at a fixed model size, with a small advantage relative to W3A3. Notably, W3A3 CAGE is \emph{Pareto-superior} to W4A4 QuEST!

\begin{figure*}[h!]
\centering

    \centering
    \includegraphics[width=0.9\linewidth]{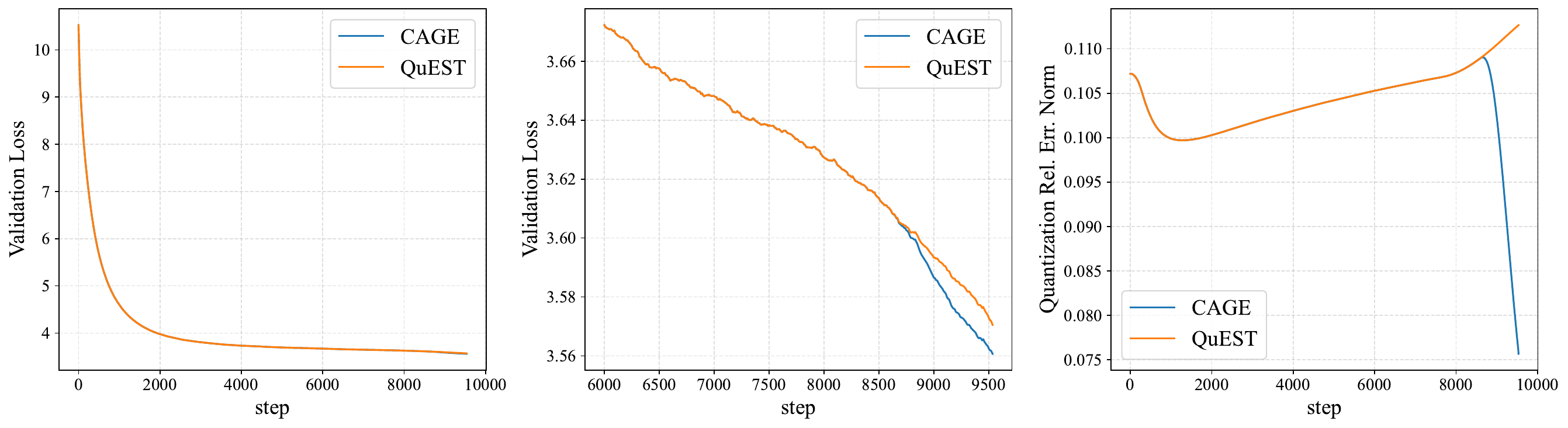}
    \caption{Illustration of the validation loss for training the 50M Llama model in W4A4 using the QuEST baseline end-to-end (left), focused on the last fraction of steps (middle), and in terms of quantization error (regularizer) values (right). 
    Observe that, while the overall training behavior is similar, CAGE significantly reduces loss for the quantized model in the final part of training.}
    \label{fig:loss_and_regularizer}

\end{figure*}

\paragraph{Loss vs. Regularizer.} 
In Figure~\ref{fig:loss_and_regularizer}, we visualize the evolution of the loss in parallel to the quantization error for pre-training the 50M Llama model in W4A4. 
We observe that, in the case of the baseline, there is a clear increase in quantization error as we continue training; at the same time, there is a sharp, consistent decrease in quantization error once the regularization starts, proportional to the value of the $\lambda$ parameter. This confirms the predicted behavior of CAGE, as well as the improvements left ``on the table'' by the standard QAT baseline.

\paragraph{Universality Across Optimizers.} To verify that gains produced by CAGE are independent of the optimizer update, we conduct a series of additional experiments using alternative optimizers to AdamW: Muon \cite{muon}, Shampoo \cite{shampoo}, and SOAP \cite{soap}. 

We train 50M models from the OLMo2 family \cite{olmo2} on the ClimbMix dataset \cite{climb_dataset} for $\frac{D}{N} = 100$ training tokens per parameter. OLMo2 models use no biases, employ rotary positional embeddings \cite{rope}, RMSNorm~\cite{rmsnorm}, and reordered pre-normalization~\cite{swin,olmo2}. For each optimizer, we first tune the hyperparameters, and then we match the training setup and weights initialization to compare training curves with and without employing CAGE method. We use CAGE hyperparameters $s = 0.9$, $\lambda=5$. 

As shown in Figure \ref{fig:optimizers-loss-curves}, CAGE consistently reduces validation loss for all examined optimizers without any additional hyperparameter tuning. Since improvements introduced by CAGE do not depend on the choice of optimizer, we conclude that it can be regarded as \emph{an optimizer-agnostic  technique}.

\paragraph{Precision Scaling Law.}
To compare methods beyond pointwise losses, we fit a separable data/model/precision-scaling law~\cite{kumar2024scaling, frantar2025compressionscaling} to the validation loss. Following  references \cite{kumar2024scaling,panferov2025quest,hoffmann2022trainingcomputeoptimallargelanguage, frantar2025compressionscaling}, we fit a law of the form:
\[
\mathcal{L}(N,D,P)\;=\;\frac{A}{\left(N\cdot \mathrm{eff}(P)\right)^{\alpha}}\;+\;\frac{B}{D^{\beta}}\;+\;E,
\]
where \(N\) is parameter count, \(D\) is the seen token count, and \(P\) denotes the bit budget. The factor \(\mathrm{eff}(P)\) captures the \emph{effective capacity} penalty due to quantization, where the capacity of standard precision is \(\mathrm{eff}(\text{FP}) = 1\).

To fit the law, we jointly fit \(A,\alpha,B,\beta,E\) \emph{shared} (jointly) across methods and learn a separate \(\mathrm{eff}(P)\) per method/bit-width via nonlinear least squares on the grid of \((N,D)\) we trained (see details in Appendix \ref{app:hyperparameters}). We regularize with weak log-priors to maintain shared values for the parameters \(\alpha,\beta>0\) across precisions. 

\paragraph{Improved Parameter Efficiency.} The obtained  \(\mathrm{eff}(P)\), normalized relative to standard 16-bit precision by multiplying by $16/P$, are plotted in Figure~\ref{fig:eff_comparison}. We observe that \methodname{} consistently increases \(\mathrm{eff}(P)\) vs. QuEST (and the other baselines), indicating improved \emph{effective} capacity for a given parameter budget. Specifically, we find that CAGE improves parameter efficiency by more than 10\% at 4-bit precision, and 20\% at 2-bit precision, relative to the prior state-of-the-art method. We also observe that 4-bit precision appears to be the ``optimal'' bit-width, although only with a small relative advantage relative to 3-bit when using CAGE.

\paragraph{Comparison to LOTION.}
We also compare \methodname{} to LOTION. At the time of writing, no official implementation of LOTION was publicly available, so we reimplemented the method based on the description in the paper. We performed an extensive sweep over the regularization coefficient and found that the best-performing setting for LOTION corresponds to a relatively large value (in the order of \(4\times 10^{3}\)). Since LOTION assumes weight-only quantization and is sensitive to saturation, we used an AbsMax weight quantizer (W4A16) in this experiment, without activation quantization.

\begin{figure}[h]
\centering

    \centering
    \includegraphics[width=0.75\linewidth]{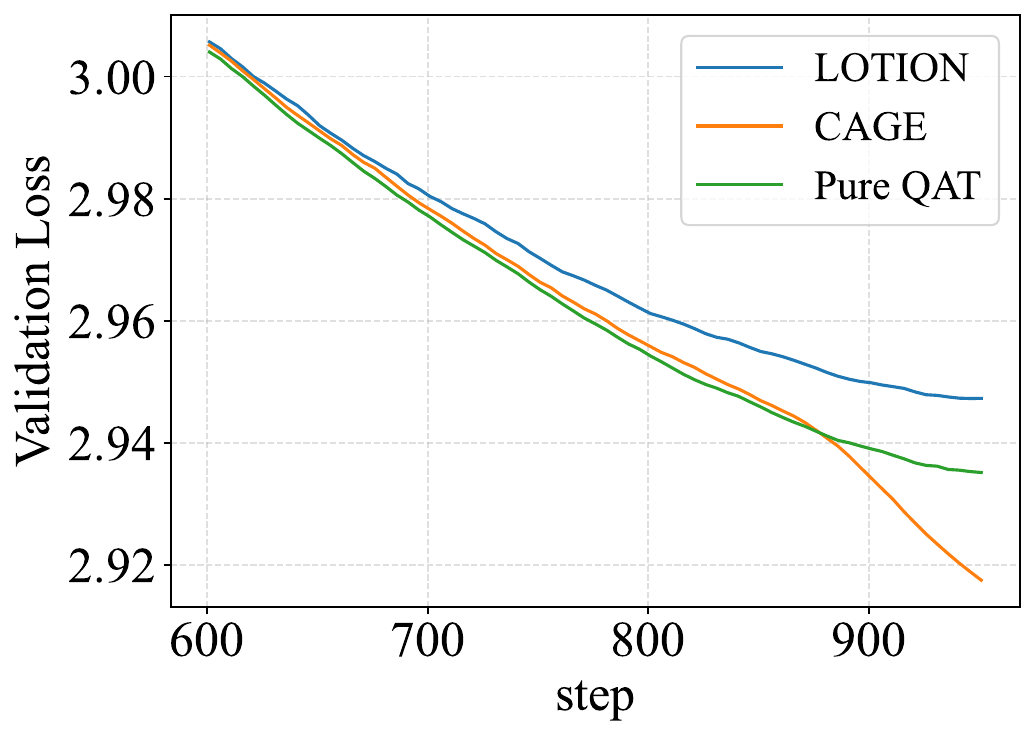}
    \caption{Pretraining loss comparison between CAGE and LOTION on a 100M-parameter Llama-style model trained on C4 under weight-only QAT with AbsMax W4A16 quantization. Both methods follow a similar trajectory early in training, but CAGE continues to improve and attains a substantially lower final loss, while LOTION plateaus.}
    \label{fig:cage_vs_lotion_loss}

\end{figure}

The comparison was carried out in our standard pretraining setup: a 100M-parameter Llama-style model trained on C4 with QAT. As shown in Figure~\ref{fig:cage_vs_lotion_loss}, both methods track a similar loss curve in the early stages of training, but \methodname{} achieves a clearly lower final validation loss, while LOTION plateaus. We note that we will re-run this comparison once an official LOTION implementation becomes available, to rule out discrepancies due to reimplementation.

\paragraph{Training Overheads.} A key practical property of \methodname{} is that it introduces negligible computational relative to the QAT baseline. The only additional operations are one extra quantization call to compute the residual $e_t = x_t - Q(x_t)$ and a small number of elementwise operations in the optimizer step. Since the dominant cost in the LLMs arise from matrix multiplications, the end-to-end overhead of these element-wise operations are small.  We measure wall-clock iteration time on a single H100 for 100M and 430M models under W4A4. Without Hadamard transforms in quantization operations, QuEST and CAGE have statistically indistinguishable iteration times ($101.6\pm1.7$ vs. $101.1\pm1.8$~ms/iter at 100M; $282.7\pm3.1$ vs. $283.1\pm3.0$~ms/iter at 430M). With HT enabled, both methods incur additional cost from the Walsh--Hadamard transform (no special fused kernels used), and CAGE remains comparable to QuEST. Full measurements are reported in Appendix~\ref{app:overheads}. Peak GPU memory is unchanged, as CAGE stores no additional persistent state beyond the elementwise residual computed on the fly.

%% file: discussion.tex
We introduced CAGE (Curvature-Aware Gradient Estimation), a novel method for quantization-aware training (QAT) designed to mitigate the accuracy degradation inherent in low-bit quantization. CAGE addresses the limitations of the straight-through estimator (STE) by augmenting the gradient with a principled, curvature-aware correction term. This approach is rooted in a reformulation of QAT as a multi-objective optimization problem, balancing the minimization of the task loss with the satisfaction of quantization constraints.

The key theoretical contribution is the definition of Pareto-optimal solutions for quantized optimization. Moreover, CAGE offers strong ergodic convergence guarantees to a Pareto-optimal point in the smooth non-convex setting, providing theoretical grounding often lacking in STE heuristics. While the approach is optimizer-agnostic, we developed a highly efficient, decoupled implementation on top of AdamW that leverages existing optimizer statistics for stability and performance. Our formulation also generalizes the concurrent work of~\cite{kwun2025lotion}.

Empirical validation on both pre-training and post-training QAT demonstrates the effectiveness of CAGE through extensive experiments involving the pre-training of Llama-style models up to 800M parameters. CAGE consistently outperformed strong baselines that employ modern outlier-mitigation techniques. Specifically, scaling law analysis confirms that CAGE improves the effective capacity of models compared to existing QAT methods.

The good performance of CAGE indicates that  incorporating local curvature information and regularization are useful for QAT performance. Future work may explore the application of CAGE in ultra-low-bit regimes and its integration with other compression techniques, such as sparsification and vector quantization.

%% file: appendix.tex
\section{Convergence Analysis: Proof of Theorem \ref{thm:main-pareto}}

    Let $F(x) = f(x) + \lambda\phi(x)$ be the overall regularized loss with smoothness constant $L =  L + \lambda L_{\phi}$. Then, due to smoothness inequality, we have
    \begin{eqnarray*}
        F(x_{t+1})
        &\le& F(x_t) + \langle \nabla F(x_t), x_{t+1}-x_t \rangle + \frac{L}{2}\|x_{t+1}-x_t\|^2 \\
        &=& F(x_t) - \alpha\langle \nabla f(x_t) + \lambda(x_t-Q(x_t)), \widetilde{\nabla} f(x_t) + \lambda(x_t-Q(x_t)) \rangle \\
        &&\quad\qquad +\; \frac{L \alpha^2}{2}\|\widetilde{\nabla} f(x_t) + \lambda(x_t-Q(x_t))\|^2.
    \end{eqnarray*}
    Applying conditional expectation $\E_t[\cdot] = \E[\cdot\mid x_t]$, using unbiasedness of stochastic gradient and boundedness of variance, we get
    \begin{eqnarray*}
        \E_t[ F(x_{t+1}) - F(x_t)]
        &\le& - \alpha \|\nabla f(x_t) + \lambda(x_t-Q(x_t))\|^2+ \frac{L \alpha^2}{2} \E_t\left[\|\widetilde{\nabla} f(x_t) + \lambda(x_t-Q(x_t))\|^2\right] \\
        &=& - \alpha \|\nabla_{\rm\lambda P} f(Q(x_t))\|^2 + \frac{L \alpha^2}{2} \left(\|\nabla_{\rm\lambda P} f(Q(x_t))\|^2 + \sigma^2 \right) \\
        &=& - \alpha\left( 1 - \frac{L\alpha}{2} \right) \|\nabla_{\rm\lambda P} f(Q(x_t))\|^2 + \frac{L \alpha^2}{2} \sigma^2 \\
        &\le& - \frac{\alpha}{2} \|\nabla_{\rm\lambda P} f(Q(x_t))\|^2 + \frac{L \alpha^2}{2} \sigma^2,
    \end{eqnarray*}
    where we enforced $\alpha \le \frac{1}{L}$ step-size restriction. Applying full expectation and summing the obtained inequalities from $t=0,\dots,T-1$, we get
    \begin{eqnarray*}
    \E \|\nabla_{\rm\lambda P} f(Q(\hat x))\|^2
    &\le& \frac{1}{T}\sum_{t=0}^{T-1} \E \|\nabla_{\rm\lambda P} f(Q(x_t))\|^2
    \le \frac{2(F(x_0) - F(x_T))}{\alpha T} + \alpha L \sigma^2 \\
    &=& \frac{2(f(x_0) - f(x_T))}{\alpha T} + \frac{2\lambda(\phi(x_0) - \phi(x_T))}{\alpha T} + \alpha L \sigma^2 \\
    &\le& \frac{1}{\sqrt{T}} \left( 2(f(x_0) - f^*) + 2\lambda(\phi(x_0) - \phi(x_T)) + L \sigma^2 \right) \max\left(1, \frac{L}{\sqrt{T}}\right) \\
    \end{eqnarray*}
    with $\alpha = \min(\frac{1}{L}, \frac{1}{\sqrt{T}})$. It remains to bound the term $\phi(x_0) - \phi(x_T)$ stemming from the quantization.

    Due to the assumption \ref{asm:quant-field} on quantization we can represent the scalar function $\phi$ as path-independent line integral:
    $$
    \phi(x) = \int_{x_0}^x (I-Q) \cdot d{\bf r},
    $$
    where $I$ is the identity map, i.e. $I(x)=x$ for any $x\in\R^d$. Since $I-Q$ is a gradient field (i.e., $x - Q(x) = \nabla \phi(x)$), the line integral above does not depend on how the endpoints $x_0$ and $x$ are connected. Therefore, we choose the direct path $r(t) = x_0 + (x-x_0)t$ for $t\in[0,1]$ and simplify the integral into usual Riemannian integral as
    $$
    \phi(x) = \int_{x_0}^x (I-Q) \cdot d{\bf r} = \int_0^1 \langle r(t) - Q(r(t)), r'(t) \rangle\, dt.
    $$
    Using this relation, we can bound the term as follows
    \begin{eqnarray*}
        \phi(x_0) - \phi(x_T)
        &\le& \left| \int_0^1 \langle r(t) - Q(r(t)), r'(t) \rangle\, dt \right| \\
        &\le& \int_0^1 \|r(t) - Q(r(t))\| \cdot \|r'(t)\| \; dt
        \le \max_{x\in[x_0,x_T]} \|x-Q(x)\| \cdot \|x_0-x_T\|.
    \end{eqnarray*}

\section{Hyperparameters and Reproducibility}
\label{app:hyperparameters}

\begin{table}[t]
\centering
\caption{Hyperparameters for Llama-style pretraining runs (per model size). Token budgets follow a 100 tokens-per-parameter (TPP) rule.}
\label{tab:llama_pretrain_hparams}
\begin{tabular}{lcccccc}
\toprule
Model ($N$) & Layers & Heads & $d_{\mathrm{model}}$ & Base LR & Tokens ($D$)  \\
\midrule
30M  & 6  & 5  & 640  & $1.2\times 10^{-3}$   & $3$B     \\
50M  & 7  & 6  & 768  & $1.2\times 10^{-3}$   & $5$B    \\
100M & 8  & 8  & 1024 & $6.0\times 10^{-4}$   & $10$B    \\
200M & 10 & 10 & 1280 & $3.0\times 10^{-4}$   & $20$B    \\
430M & 13 & 13 & 1664 & $1.5\times 10^{-4}$   & $43$B    \\
800M & 16 & 16 & 2048 & $7.5\times 10^{-5}$   & $80$B    \\
\bottomrule
\end{tabular}

\vspace{0.6em}
\small

\end{table}

\textbf{Shared settings.} Sequence length $L=512$; batch size $64$ with gradient accumulation $8$ (effective tokens/step fixed across sizes). Optimizer AdamW with $\beta_1 = 0.9$, $\beta_2 = 0.95$, $\varepsilon = 10^{-8}$, weight decay $0.1$, gradient clip $1.0$. Cosine LR schedule with warmup $10\%$ of total steps. FP32 master weights and optimizer states; bfloat16 compute. Hardware: $8\times$ H100 (NCCL, \texttt{compile} enabled).

\textbf{Quantization.} QuEST row-wise Hadamard quantizer for weights \emph{and} activations; bit-widths $b\in\{2,3,4\}$ (symmetric, MSE-optimal Gaussian clipping). Trust-masked STE in transform domain.

\textbf{CAGE settings.} Throughout the experiments, unless otherwise stated, we use the decoupled variant, with silence ratio $s=0.9$, and regularization parameter $\lambda=2.0$.

\begin{figure*}[h]
\centering
\includegraphics[width=0.45\columnwidth]{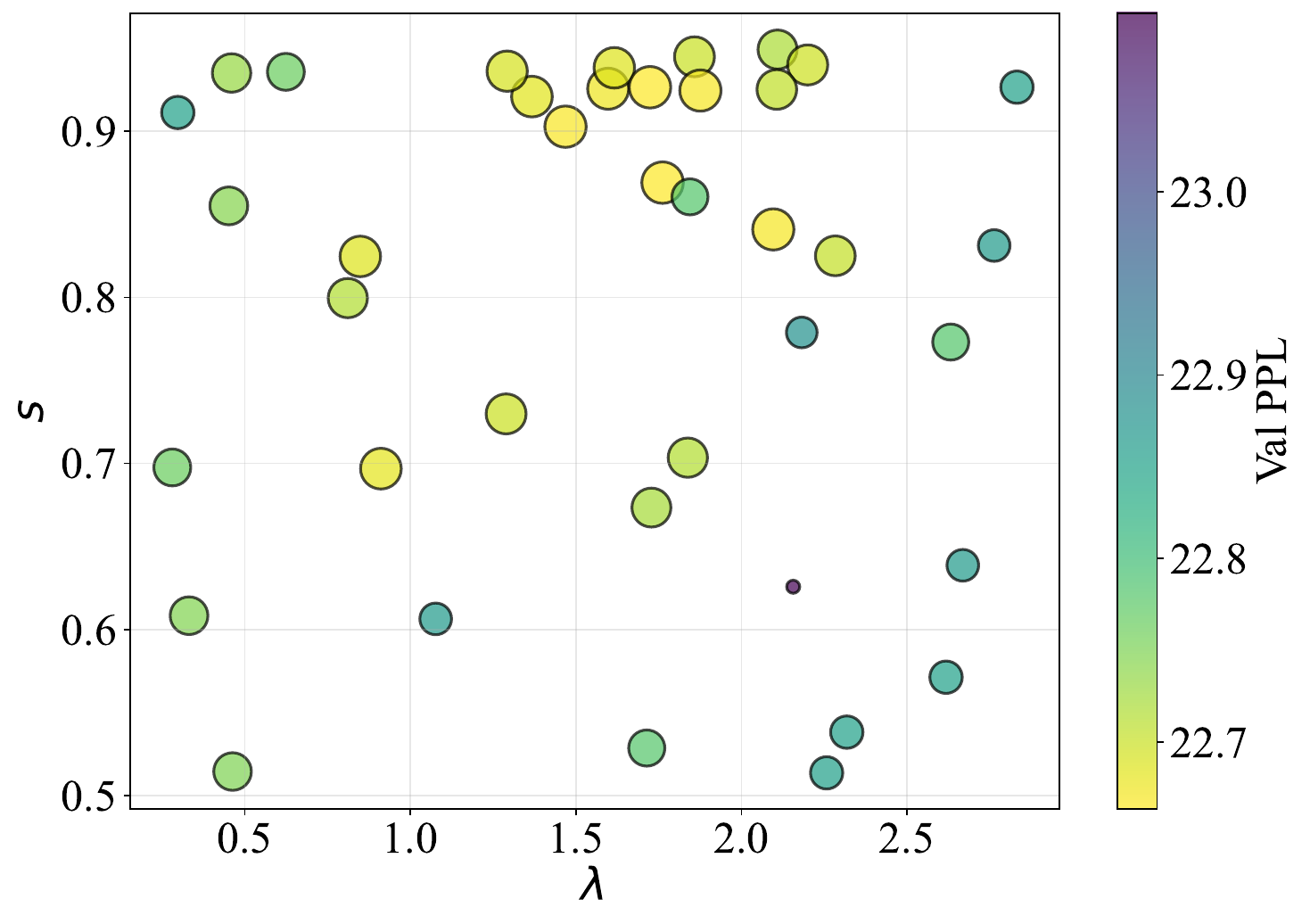}
\caption{Validation perplexity for CAGE under different choices of silence ratio \(s\) and regularization coefficient \(\lambda\) on the 50M Llama model with W4A4 quantization.}
\label{fig:sweep_s_lambda}
\end{figure*}

\section{Ablation of CAGE Hyperparameters}
\label{app:cage_ablation}
We conduct a two-dimensional sweep over the CAGE hyperparameters; the silence ratio \(s\) and the regularization coefficient \(\lambda\). Figure~\ref{fig:sweep_s_lambda} reports validation perplexity for runs on a 50M Llama model trained on C4 in the W4A4 setting. We find that \(s \in [0.8, 0.95]\) and \(\lambda \in [1, 2.5]\) form a reliable operating region. Within this range, CAGE consistently improves over the baseline.

\section{Decoupled vs coupled CAGE}
\label{app:de_vs_coupled_cage}

We extend Table~\ref{tab:pretrain_w4a4} with a direct comparison between the coupled and
decoupled formulations of CAGE. Table~\ref{tab:de_vs_coupled_cage} shows that
$\mathrm{CAGE}_C$ (coupled) and $\mathrm{CAGE}_D$ (decoupled) achieve nearly identical
final validation perplexity across model sizes, both with and without Hadamard transforms (HT),
and consistently outperform the QuEST baseline.

This behavior is expected: $\mathrm{CAGE}_C$ follows directly from the Pareto-stationary
condition in our analysis, while $\mathrm{CAGE}_D$ is a practical variant that decouples the
correction from curvature/variance scaling to improve compatibility with a broader set of
optimizers and training stacks. The empirical equivalence under AdamW indicates that the primary
gains come from the Pareto-derived correction direction, and the decoupled formulation preserves
these gains while simplifying integration.

\begin{table*}[h!]
\centering
\caption{Pretraining results (final validation perplexity; lower is better) for W4A4 across model sizes. $\text{CAGE}_C$ is the coupled variant and $\text{CAGE}_D$ is the decoupled variant.}
\vspace{0.5em}
\label{tab:de_vs_coupled_cage}
\begin{tabular}{lcccccc|c}
\toprule
Method & 30M & 50M & 100M  \\
\midrule
$\text{CAGE}_{D}$ {\small + HT }        & 26.277 & 22.747 & 18.944  \\
$\text{CAGE}_{C}$ {\small + HT }        & 26.271 & 22.752 & 18.948 \\
QuEST {\small + HT }       & 26.475 & 23.062 & 19.123  \\
\midrule
$\text{CAGE}_{D}$ {\small (no HT)}     & 27.287 & 23.781 & 19.630  \\
$\text{CAGE}_{C}$ {\small (no HT)}     & 27.285 & 23.786 & 19.625  \\
QuEST {\small (no HT)}    & 27.401 & 23.991 & 19.799 \\
\midrule
BF16 & 24.715 & 21.491 & 17.923  \\
\bottomrule
\end{tabular}
\end{table*}

\section{Algorithm Overheads}

\label{app:overheads}

We quantify the runtime (and qualitative memory) overhead of CAGE relative to the QuEST baseline
under identical training configurations: same model, sequence length, batch size, optimizer,
quantization settings (W4A4), and hardware (single NVIDIA H100). CAGE adds one extra computation of
the quantization residual $e_t = x_t - Q(x_t)$ (i.e., one additional quantization call) and a small
number of elementwise operations (subtraction and an extra update add). Since the dominant cost in
LLM training is matrix multiplications (forward/backward), we expect the end-to-end overhead to be
small.

\paragraph{Implementation note.}
All quantization operations in these measurements are implemented in PyTorch for simplicity and
reproducibility (i.e., not using a hand-tuned low-bit kernel). While the absolute iteration times
therefore do not reflect fully optimized production implementations, the comparison between QuEST
and CAGE is fair because both methods use the same quantization implementation and training stack.
When Hadamard transforms (HT) are enabled, the runtime is additionally influenced by the (naive)
Walsh--Hadamard transform used in our implementation; an optimized WHT implementation would reduce
this component for both methods.

\paragraph{Measurement protocol.}
We measure wall-clock iteration time using CUDA-synchronized timing over a steady-state window of
$N$ consecutive iterations after an initial warmup period (excluding compilation and startup
transients), and report mean $\pm$ standard deviation. Measurements use sequence length 2048 and
batch size 16. 

\paragraph{Results.}
Table~\ref{tab:overheads} reports iteration time for two model sizes. Without HT, QuEST and CAGE
have statistically indistinguishable iteration times (differences are within run-to-run variance).
With HT enabled, iteration times are more sensitive to the WHT implementation; we observe that
CAGE remains comparable to QuEST, with modest variability across model sizes. Overall, these
results support that CAGE introduces negligible end-to-end overhead relative to the baseline in
our training setup.

\begin{table}[h!]
\centering
\caption{Runtime per iteration (ms; mean $\pm$ std) for W4A4 training on a single H100 (sequence length 2048, batch size 16).}
\vspace{0.5em}
\label{tab:overheads}
\begin{tabular}{lcc}
\toprule
Method & 100M (ms/iter) & 430M (ms/iter) \\
\midrule
QuEST (no HT)    & $101.6 \pm 1.7$ & $282.7 \pm 3.1$ \\
CAGE$_D$ (no HT) & $101.1 \pm 1.8$ & $283.1 \pm 3.0$ \\
\midrule
QuEST (+HT)      & $117.5 \pm 1.8$ & $286.6 \pm 2.3$ \\
CAGE$_D$ (+HT)   & $105.8 \pm 1.2$ & $316.1 \pm 3.7$ \\
\bottomrule
\end{tabular}
\end{table}

\section{Long-context evaluation}
\label{app:ruler}

We evaluate long-context behavior using RULER~\cite{hsieh2024ruler} on a Llama 3.1-8B model fine-tuned with QAT.

\paragraph{Setup.}
We fine-tune Llama 3.1-8B on TULU-SFT under W4A16 quantization and compare CAGE against QuEST and
additional common post-training quantization baselines (RTN, GPTQ). We report the average RULER
score (higher is better). All evaluated models share the same base architecture and are compared
under the same evaluation protocol.

\paragraph{Results.}
Table~\ref{tab:ruler} shows that CAGE improves long-context performance over the QuEST baseline
under W4A16, and substantially outperforms RTN and GPTQ in this setting. While quantization
reduces performance relative to the full-precision base model, CAGE narrows this gap compared to
prior QAT baselines, indicating that the benefits of our correction extend to long-context
behavior.

\begin{table}[h!]
\centering
\caption{RULER average score (higher is better) for Llama 3.1-8B under W4A16 after fine-tuning on TULU-SFT.}
\vspace{0.4em}
\label{tab:ruler}
\begin{tabular}{lc}
\toprule
Model & Avg. RULER score \\
\midrule
Llama 3.1-8B (base, full precision) & 88.3 \\
Llama 3.1-8B (CAGE, W4A16) & 73.2 \\
Llama 3.1-8B (QuEST, W4A16) & 68.7 \\
Llama 3.1-8B (GPTQ, W4A16) & 65.1 \\
Llama 3.1-8B (RTN, W4A16) & 41.5 \\
\bottomrule
\end{tabular}
\end{table}

\section{Formal Justification for $\lambda$ Scheduler}
\label{app:lambda_schedule}

Recall that the decoupled CAGE update applies the instantaneous
quantization error
\begin{equation}
e_t := x_t - Q(x_t) \qquad 
\end{equation}
as a post-step correction:
\begin{equation}
\tilde x_{t+1} = \textsc{OptStep}(x_t; g_t), \qquad
x_{t+1} = \tilde x_{t+1} - \alpha \lambda_t e_t.
\label{eq:decoupled_update_app}
\end{equation}
Empirically (and as stated in \S3.3), turning on the correction too early can over-constrain the dynamics; we therefore use
\begin{equation}
r_t := t/T, \qquad
\lambda_t =
\begin{cases}
0, & r_t \le s, \\
\lambda \cdot \dfrac{r_t - s}{1-s}, & r_t > s,
\end{cases}
\label{eq:warmup_schedule_app}
\end{equation}
for a silence ratio $s\in[0,1)$.

We provide a complementary formal perspective to show that during the silence period, applying CAGE is ineffective and may also undermine training stability.

\subsection{Quantization as a random walk}
\label{app:cell_switching}

Let $\mathcal{Q}$ be the (discrete) quantized support and $Q:\mathbb{R}^d\to \mathcal{Q}$ the quantizer.
For any $q\in\mathcal{Q}$ define the cell
$\mathcal{C}(q) := \{x\in\mathbb{R}^d: Q(x)=q\}$.
Let $\{\mathcal{F}_t\}_{t\ge 0}$ be the filtration generated by the algorithm up to iteration $t$.
Define the \emph{cell-switch} times
\begin{equation}
\tau_0 := 0, \qquad
\tau_{k+1} := \min\{t>\tau_k:\ Q(x_t)\neq Q(x_{\tau_k})\}.
\end{equation}
Within a visit $t\in[\tau_k,\tau_{k+1})$ the quantized anchor $Q(x_t)$ is constant.

The residuals $e_t$ are correlated inside a cell (since the same anchor $Q(x_t)$ is reused), but they can "reset" when the
iterate crosses to a new cell. We capture this by grouping residuals per-visit:
\begin{equation}
E_k := \sum_{t=\tau_k}^{\tau_{k+1}-1} e_t.
\label{eq:block_sum}
\end{equation}

\subsection{Assumptions}
The following conditions formalize the intuition used in the rebuttal.

\paragraph{Assumption A (bounded residual).}
There exists $B>0$ such that $\|e_t\|\le B$ for all $t$ almost surely.
For standard uniform (or clipped) quantizers this holds since each coordinate error is bounded by half a bin width.

\paragraph{Assumption B (bounded dwell time during exploration).}
During the \emph{exploration phase} (early training, larger effective steps), cell visits are not extremely long:
there exists $L_{\max}\ge 1$ such that for all visits occurring in the exploration phase,
\begin{equation}
\tau_{k+1}-\tau_k \le L_{\max}.
\label{eq:dwell_time}
\end{equation}
This reflects the empirical regime where $x_t$ traverses many nearby quantization points.

\paragraph{Assumption C (reset symmetry / mean-zero block residual).}
During the exploration phase, the per-visit residual block satisfies an (approximate) martingale-difference property:
\begin{equation}
\mathbb{E}\!\left[E_k \mid \mathcal{F}_{\tau_k}\right] = 0
\quad\text{or more generally}\quad
\left\|\mathbb{E}\!\left[E_k \mid \mathcal{F}_{\tau_k}\right]\right\|\le \mu
\label{eq:reset_symmetry}
\end{equation}
for some small bias $\mu\ge 0$.
Intuitively, when the iterate enters a new cell, the offset from the cell center is roughly symmetric due to stochasticity, so the visit-integrated residual has a small conditional mean.

\subsection{Cancellation bound}
Assumptions A--C yield a quantitative "random-walk cancellation" statement.

\begin{lemma}[Bounded block increments]
\label{lem:bounded_blocks}
Under Assumptions A--B, for each exploration-phase visit,
\begin{equation}
\|E_k\| \le \sum_{t=\tau_k}^{\tau_{k+1}-1}\|e_t\| \le B(\tau_{k+1}-\tau_k)\le BL_{\max}.
\end{equation}
\end{lemma}

\begin{lemma}[concentration across visits]
\label{lem:azuma_visits}
Assume $\mu=0$ in \eqref{eq:reset_symmetry}. Then for any unit vector $u\in\mathbb{S}^{d-1}$ and any $\varepsilon>0$, by Azuma concentration bound,
\begin{equation}
\mathbb{P}\!\left(\left|u^\top \sum_{k=0}^{K-1} E_k\right|\ge \varepsilon\right)
\;\le\;
2\exp\!\left(-\frac{\varepsilon^2}{2K(BL_{\max})^2}\right),
\end{equation}
\end{lemma}
And consequently $\left\|\sum_{k=0}^{K-1}E_k\right\| = O_{\mathbb{P}}(BL_{\max}\sqrt{K})$ via a standard $\varepsilon$-net argument.

Let the exploration phase span $T_{\mathrm{exp}}$ iterations and $K(T_{\mathrm{exp}})$ visits. When switching is frequent,
$K(T_{\mathrm{exp}})=\Theta(T_{\mathrm{exp}})$ and Lemma~\ref{lem:azuma_visits} implies
\begin{equation}
\left\|\frac{1}{T_{\mathrm{exp}}}\sum_{t=0}^{T_{\mathrm{exp}}-1} e_t\right\|
=
\left\|\frac{1}{T_{\mathrm{exp}}}\sum_{k=0}^{K(T_{\mathrm{exp}})-1} E_k\right\|
=
O_{\mathbb{P}}\!\left(\frac{BL_{\max}}{\sqrt{T_{\mathrm{exp}}}}\right).
\label{eq:avg_residual_small}
\end{equation}
Thus, \emph{early} residuals behave like bounded, approximately mean-zero perturbations: they cancel in aggregate but have
$\sqrt{T}$-scale fluctuations.

\subsection{Why large constant $\lambda$ early can be harmful}
If $\lambda_t\equiv\lambda$ were active from the start, the cumulative correction displacement over the exploration phase is
\begin{equation}
\left\|\sum_{t=0}^{T_{\mathrm{exp}}-1}\alpha\lambda e_t\right\|
=
\alpha\lambda \left\|\sum_{k=0}^{K(T_{\mathrm{exp}})-1} E_k\right\|
=
O_{\mathbb{P}}\!\left(\alpha\lambda BL_{\max}\sqrt{K(T_{\mathrm{exp}})}\right).
\label{eq:cumulative_correction_noise}
\end{equation}
Even if the mean is near zero, the \emph{variance} grows with time like a random walk; a large early $\lambda$ therefore injects
a nontrivial noise-like forcing term into the dynamics, potentially destabilizing training before the iterate has settled into a basin.
This motivates \emph{silencing} the correction during early exploration (setting $\lambda_t\simeq 0$ for $t\le sT$).

\subsection{Why $\lambda_t$ should increase late}
Near convergence, the iterate typically oscillates among a small set of adjacent cells for long periods,
so the "reset" condition \eqref{eq:reset_symmetry} weakens. In the extreme, if $Q(x_t)$ is constant for
$t\in[t_0,t_0+M)$, then $e_t$ no longer cancels and
$\left\|\sum_{t=t_0}^{t_0+M-1} e_t\right\|$ can scale as $\Theta(M)$.
In this regime the correction is \emph{coherent} and effectively pushes $x_t$ toward the quantized support, which is precisely when
enforcing Pareto-stationarity becomes beneficial. Hence $\lambda_t$ should be \emph{larger} late in training.

\subsection{Dynamic quantizers where $L \approx 1$}
\label{app:dynamic_quantizers}

\paragraph{Code-based visits.}
Define visits by the \emph{integer code}, not by the reconstructed value.

Write the dynamic quantizer as
\[
c_t := C(x_t, s_t)
:= \mathrm{clip}\!\left(
    \mathrm{round}\!\left(\frac{x_t}{s_t}\right),
    [-M, M]
\right)
\in \mathbb{Z}^d,
\qquad
q_t := Q_{s_t}(x_t) = s_t\, c_t .
\]

The residual is then
\[
e_t = x_t - q_t
= x_t - s_t c_t
= s_t\!\left(\frac{x_t}{s_t} - c_t\right)
=: s_t\, r_t ,
\]
where
\[
r_t := y_t - \mathrm{round}(y_t),
\qquad
y_t := \frac{x_t}{s_t}.
\]

\paragraph{Visit definition.}
Define "visits" by \emph{code stability}:
\[
\tau_{k+1}
:= \min \bigl\{ t > \tau_k : c_t \neq c_{\tau_k} \bigr\}.
\]

The sequence $(c_t)$ can remain constant for many steps even if $(s_t)$ changes at every step.
Thus, the "visit time $=1$" issue disappears.

\paragraph{Assumption.}
A defensible assumption is the following:

\begin{quote}
During the exploration phase, conditional on $s_t$ (or more generally on $\mathcal{F}_{t-1}$),
the normalized fractional part
\[
r_t = y_t - \mathrm{round}(y_t)
\]
is approximately symmetric around $0$ and mixes across time. Consequently,
\[
\mathbb{E}\!\left[ r_t \mid \mathcal{F}_{t-1} \right] \approx 0 .
\]
\end{quote}

It follows immediately that
\[
\mathbb{E}\!\left[ e_t \mid \mathcal{F}_{t-1} \right]
= \mathbb{E}\!\left[ s_t r_t \mid \mathcal{F}_{t-1} \right]
= s_t\, \mathbb{E}\!\left[ r_t \mid \mathcal{F}_{t-1} \right]
\approx 0 .
\]

\paragraph{Block sums over code visits.}
For block sums over \emph{code visits},
\[
E_k
:= \sum_{t=\tau_k}^{\tau_{k+1}-1} e_t
= \sum_{t=\tau_k}^{\tau_{k+1}-1} s_t r_t ,
\]
the same martingale and concentration arguments as before apply.
Now, however, the "sticking" correlation arises from the code remaining constant,
which is the appropriate notion of a visit.